# Enhancing Accuracy and Maintainability in Nuclear Plant Data Retrieval: A Function-Calling LLM Approach Over NL-to-SQL

**Mishca de Costa[1,2], Muhammad Anwar[1,2], Dave Mercier[1,2], Mark Randall[1], and Issam Hammad[2]**

[1] Data Analytics and AI, Digital Technology and Services, Ontario Power Generation, Pickering, Ontario, Canada

[2] Department of Engineering Mathematics and Internetworking, Faculty of Engineering, Dalhousie University, Halifax, Nova Scotia, Canada

mishca.decosta@opg.com

## Abstract

Retrieving operational data from nuclear power plants requires exceptional accuracy and transparency due to the criticality of the decisions it supports. Traditionally, natural language to SQL (NL-to-SQL) approaches have been explored for querying such data. While NL-to-SQL promises ease of use, it poses significant risks: end-users cannot easily validate generated SQL queries, and legacy nuclear plant databases—often complex and poorly structured—complicate query generation due to decades of incremental modifications. These challenges increase the likelihood of inaccuracies and reduce trust in the approach. In this work, we propose an alternative paradigm: leveraging function-calling large language models (LLMs) to address these challenges. Instead of directly generating SQL queries, we define a set of pre-approved, purpose-specific functions representing common use cases. Queries are processed by invoking these functions, which encapsulate validated SQL logic. This hybrid approach mitigates the risks associated with direct NL-to-SQL translations by ensuring that SQL queries are reviewed and optimized by experts before deployment. While this strategy introduces the upfront cost of developing and maintaining the function library, we demonstrate how NL-to-SQL tools can assist in the initial generation of function code, allowing experts to focus on validation rather than creation. Our study includes a performance comparison between direct NL-to-SQL generation and the proposed function-based approach, highlighting improvements in accuracy and maintainability. This work underscores the importance of balancing user accessibility with operational safety and provides a novel, actionable framework for robust data retrieval in critical systems.

## 1. Introduction

Operational data retrieval in nuclear power plants demands unparalleled accuracy and transparency due to its critical impact on decision-making. Traditional NL-to-SQL methods, while user-friendly, introduce risks when handling complex legacy databases. This paper introduces a function-calling LLM approach that leverages pre-approved, purpose-specific functions to encapsulate validated SQL queries. By combining expert review with automated NL-to-SQL tools, the proposed framework enhances both the accuracy and maintainability of data retrieval processes.

The Berkeley Function-Calling Leaderboard (BFCL) ranks LLMs based on function execution accuracy across various tasks. GPT-4 (2024) leads in performance, followed closely by Gorilla OpenFunctions-v2, Mistral, and Claude-2.1 [3]. The ToolACE-8B model, optimized with targeted fine-tuning on synthetic tool-use datasets, achieves GPT-4-level performance despite being significantly smaller [4].

Benchmarks such as HammerBench and AgentBench evaluate multi-turn API usage, exposing common LLM errors, including incorrect parameter formatting, function misselection, and difficulty maintaining context in extended interactions [5][6]. AgentBench tests LLMs on increasingly complex tool-use scenarios, revealing that smaller models, while competitive in single-turn tasks, struggle with multi-step workflows requiring memory retention [6].

GPT-4 and Gorilla OpenFunctions demonstrate superior handling of nested and parallel API calls, allowing them to complete complex workflows more efficiently. Open-source models, though improving, require specialized training to bridge this gap [4].

As the number of available tools increases, LLM decision-making degrades significantly. Although most systems do not impose a hard cap on the number of tools, studies indicate that performance declines after 20-25 functions due to cognitive overload [7]. For example, in empirical trials, reducing the number of tools from 46 to 19 notably improved function selection accuracy and response time [7].

To improve function selection accuracy, researchers have developed retrieval-based tool filtering methods that reduce the number of tools an LLM considers per query. The Gorilla system uses a document retriever to fetch only the most relevant API documentation, minimizing cognitive load and improving selection precision by preventing tool overload [2]. This approach ensures the LLM focuses on a manageable subset of tools, avoiding errors caused by excessive choices.

Future optimizations involve adaptive function selection, where LLMs dynamically adjust tool availability based on real-time context and past interactions. Other solutions propose a retrieval-augmented generation (RAG) strategy that combines retrieval, fine-tuned models, and execution-feedback learning to improve tool selection efficiency [3]. By leveraging these methods, LLMs can enhance real-world tool use while maintaining high selection accuracy and minimal errors.

## 2. Challenges

We faced several technical challenges that influenced our methodology. This section discusses these challenges, and the approaches considered to solve them.

### 2.1 Agentic/Autonomous Behaviour

The objective of our function-calling structured SQL method is to minimize autonomous behavior by providing explicit instructions to agents on the necessary steps to follow. However, there are instances where the agent must make an autonomous decision based on the user's query.

This issue mainly arises when determining whether to perform data retrieval. For example, if a user asks, "What day is it today?" or "What functions do you support?", the bot should not execute any SQL retrieval but should instead respond directly using its internal capabilities.

Fortunately, this behavior is inherently supported in function-calling large language models (LLMs). The LLM determines whether to call a function based on the user query and the instructions given in the system prompt or developer message. However, less advanced models encounter difficulties with this decision-making process and require extensive, detailed prompts to guide them on which tools to use and when. This makes maintaining the system challenging.

Meta advises that their 8 billion parameter model is only adequate for zero-shot function calling, recommending at least 70 billion parameters if both conversation and function calling need to be supported simultaneously [1]. Our own testing corroborated these findings, revealing that smaller models generally fail to concurrently handle conversation and function calling effectively. Models below 70 billion parameters typically struggle with function calling and conversational tasks, although specialized models in the 20-30 billion parameter range, such as recent coding assistants, are becoming available. A comprehensive and quantitative assessment of function-calling performance is documented in the literature review section below.

### 2.1.1 Follow Up Questions

Another area where the conversational behavior of the LLM system is critical is in handling follow-up questions. For instance, a user may request data that necessitates retrieval, but their subsequent question might simply involve formatting the data into a table, which does not require further retrieval. The LLM agent must discern this difference and avoid unnecessary retrievals. This is important not only to reduce latency and costs, but also to maintain quality. Adding more context before performing the task can also lead to hallucinations caused by distracting the LLM with irrelevant/superfluous content.

## 2.2 Technical Jargon

User queries often include terminology that is not found in natural language, which the LLM struggles to interpret. This may include acronyms like WR (Work Request) or technical jargon like "megger the motor" (using a Megger brand insulation resistance tester). Staff use such terms daily, so breaking them down for the bot isn't intuitive. The bot must handle these terms. Adding jargon definitions to the system prompt/developer message can help, but several challenges remain:

- Capturing every possible acronym or jargon term isn't feasible, though OPG is working on a comprehensive database.
- Some acronyms have multiple meanings and need context for correct interpretation. Sentences can be almost entirely acronyms, like: "find all the WRs against 056-SG2 which caused a SDNMDI or TLOR on the SG bank."
- Including too many definitions in the system prompt can distract the LLM, reducing its performance.

While placing definitions in specific tool calls can partly mitigate this, it means repeating definitions across tools, leading to maintainability issues and increasing token payload size.

## 2.3 Structured Outputs

One challenge with agentic LLM systems is ensuring that every piece of data exchanged fits a strict schema or structure. We chose JSON as our preferred format because the data has to be structured, and JSON offers a standard, human-readable way to do so. Whether it's parameters for function calls or data fetched from a database, even slight deviations from the expected structure can lead to crashes or unpredictable behavior. This risk is compounded by issues with data types—for example, the LLM might output a date in a non-standard format or return a string where an integer is expected. To address this, we rely on Pydantic models to enforce precise data types (dtypes) and validate every parameter. If the output doesn't conform to the schema, the system automatically retries the operation, often correcting the issue on subsequent attempts.

OpenAI's structured output mode further enhances reliability through a technique called constrained decoding. Constrained decoding manipulates the token generation process to ensure that outputs strictly adhere to a predefined schema. It begins by defining a schema—such as a JSON Schema—that outlines the desired structure, which is then converted into a context-free grammar (CFG) that encapsulates the rules of the output language [2]. During token generation, the model dynamically constrains its choices by considering previously generated tokens, the CFG rules, and the current position in the output structure [2]. This process includes token masking, where the list of valid tokens is used to lower the probability of invalid tokens to zero [2]. Additionally, selective generation allows the model to produce tokens only for parts of the output that require active generation—skipping boilerplate or deterministic elements [1][3]—while optimizations during preprocessing help speed up generation and reduce latency [1][3]. This constrained decoding approach ensures 100% consistency with the specified output format and is particularly effective for handling complex, nested, or recursive data structures [3].

### 2.4 Retries

Retries have been identified as an effective technique to address issues related to the structure of LLM systems. They can also be utilized to manage other failures within the LLM system. Occasionally, the LLM may call upon the incorrect agent due to contextual clues in the user's query that suggest multiple agents. For instance, a query like "how many work orders are filed against SG2" might be directed to the equipment agent instead of the work order agent, given its reference to equipment. In such cases, the equipment agent would not return any data for this query, necessitating the main agent to perform a retry. The system imposes limits on retries to prevent indefinite attempts when no suitable function is available. However, this limitation means that at least the minimum number of retries will always be executed, even when no applicable function exists.

### 2.5 Chained Queries

Sometimes the user's query may require a chained tool call. For example, say the user wanted to know how many bearings are available for a piece of equipment. We have a tool which can check how many bearings are in stock, but the user did not specify the part number, they specified the equipment number, so we must first use our equipment bill of material tool to get the list of materials, identify which part number is for the bearing, and then use that to call the stock tool to get the quantity in stock. This can be very challenging for the system as it needs to break the problem down into pieces and establish a plan. This is currently not supported in this version of our system, but we plan to add a reasoning model as the first step to make a detailed plan for the other agents to execute.

### 3. Final Methodology

In this paper we discuss the methodology used for a function calling SQL retrieval agentic workflow. This system is not fully agentic as the agents are not given full autonomy to make decisions, this is a workflow process where each agent is guided by specific rules and structure in terms of when they get called and in what order.

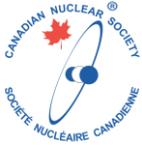

## 3.1 Detailed System Overview for Function Calling Approach

The system is designed as a multi-agent, function-calling architecture that bridges enterprise data queries to a robust SQL database. Its overall goal is to answer user questions about enterprise data (for example, work orders, maintenance schedules, catalogue IDs, etc.). The system's architecture is built upon a few key design principles:

- Query Routing via Agents: When a user submits a natural language query, the central "main agent" receives it. The main agent has access to a suite of tools or sub agents that each encapsulate a set of tools of functions for a specific domain of queries. For example, a work order sub agent will be responsible for handling the work order related SQL query functions.

- Iterative Dialogue and Retry Logic: Only the sub agents can call upon and execute functions to run SQL queries. Both the main and sub agents will maintain a history of each interaction including system prompts, the users query, tool calls/results, and its own answer. This allows sub agents to retry if they make a mistake and call an inappropriate function, or the main agent to try a different sub agent in the even it gets an inappropriate response back from the sub agent called.

- Logging and Error Handling: Most errors can be handled by the LLM agents themselves (such as determining when a retry is necessary). Each step is also logged as they happen in an SQL database including the users initial query, the tool calls/results along the way, and the final response. Any failures the LLMs cannot handle are caught and logged to the DB to ease the process of troubleshooting.

- Overall Methodology: The overall methodology is based on combining the NL reasoning power of LLMs with strict control over database interactions by limiting the SQL generation to predefined queries and only using the reasoning of the LLM to decide which interaction is appropriate. Parameter validation systems and retry logic ensure both a robust workflow while protecting against SQL injection.

A diagram of the workflow is illustrated in Figure 1.

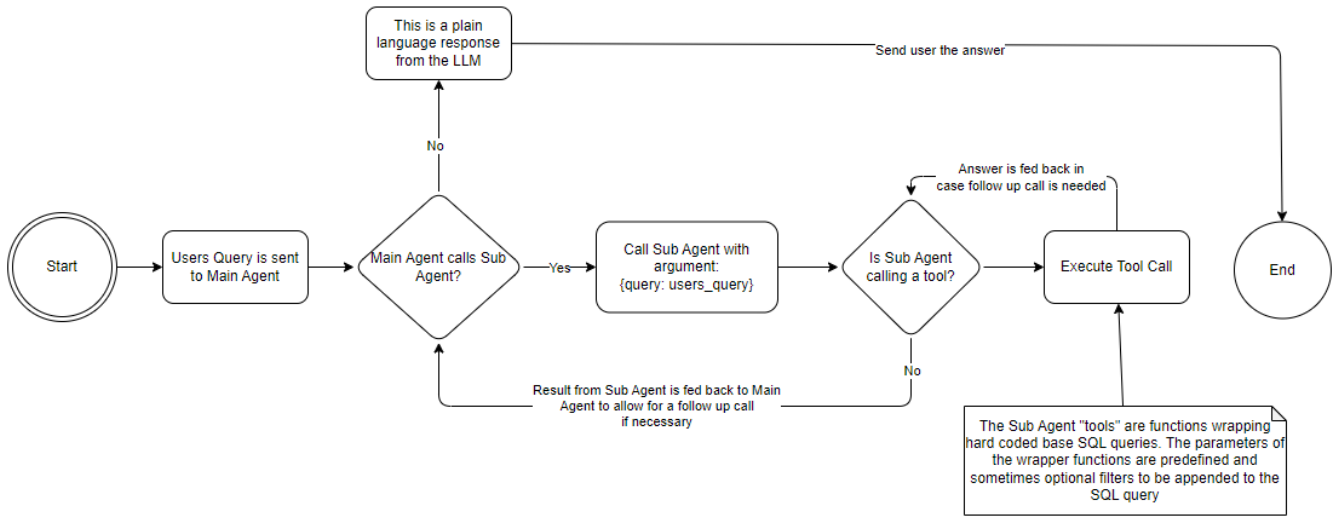

Figure 1 – Function Calling Approach Flow diagram

### 3.2 Detailed System Overview for Non-Function Calling Approach

This multi-agent system uses a blended RAG (Retrieval-Augmented Generation) approach alongside a custom NL-to-SQL query agent that creates queries based on example queries and system knowledge, without explicitly using function-calling—differing from the main solution presented previously. The system ensures accurate table and field selection, corrects mismatches, formats queries, and combines deterministic query results with the flexibility to create new queries based on its schema knowledge. Figure 2 illustrates a flowchart for the approach.

1. Receive User Query: The program starts by taking a user's natural language question (for example, "Show me all the work requests entered in by John Smith").

2. Extract Query Intent: The application calls an intent extraction agent that analyzes the query to determine the user's intent and key entities (like tables, fields, conditions, and work areas).

3. Retrieve Example Queries: Using the extracted intent as context, the system retrieves similar or relevant example queries from a vector-based search index.

   a. Decision Point: If one of the examples closely matches the intent, it is selected, and its parameters are replaced with the actual values from the query, as identified in the intent.
   b. Otherwise, the system uses its built-in knowledge of the schema and inspired by the retrieved examples to generate a new SQL query from scratch.

4. Generate Initial SQL Query: Based on the chosen path, an initial SQL query is constructed. This query also comes with an explanation of how it was built.

5. Validate SQL Query: The initial SQL query is then passed to a validation agent that cross-checks the table and field names against the system's schema (stored in a vector database). Any discrepancies are corrected so that the query conforms to the expected format and contains only valid tables and fields.

6. Execute SQL Query: The system then executes the validated SQL query, and results are retrieved and converted to JSON for further processing.

7. Generate User-Friendly Answer: Finally, an agent uses the query results to generate a concise, clear answer that directly addresses the original question.

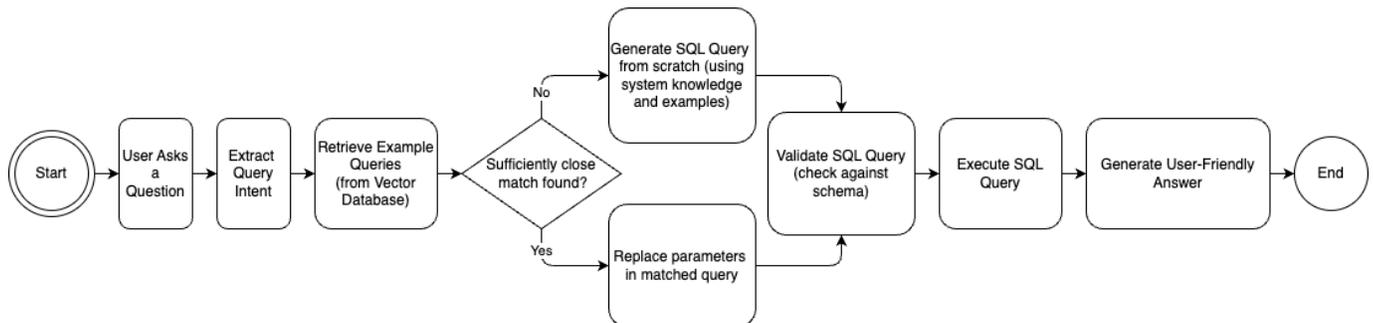

Figure 2 – Non-Function Calling Approach Flowdiagram

## 3.3 Model Selection

For this project we used OpenAI's GPT-4o model. At the time of writing, based on the analysis in the literature review, GPT-4o is the best model on the market today in terms of function calling performance. Given that GPT-4o is in the top 10 of almost every benchmark including coding, math, logic, etc, it is a strong candidate to be used for this method. By using the most powerful generalist model available we can more readily compare raw NL-to-SQL performance by using the model for both systems isolating the variability of the LLM as best we can.

## 4. Evaluation

In order to evaluate the performance of our proposed function calling methodology, we evaluate the outputs of our two models against various factors.

### 4.1 Answer Relevance (Query-Only)

This metric focuses solely on how well the response addresses the user's specific query, independent of any external context. It assesses whether the answer:

- Directly and clearly responds to the user's question.

- Includes all aspects requested in the query without adding unrelated details.

- Uses precise language and structure that aligns with the subject of the query's intent.
  For example, if the user asks for a summary of SQL query results, the answer should provide a concise summary that focuses strictly on that request.

### 4.2 Relevance (Query Plus Context)

This broader relevance measure considers both the user's query and the accompanying context (in this case, SQL query results). It evaluates whether the answer:

- Effectively integrates and leverages the provided SQL data to enrich the response.

- Aligns the response with both the explicit question and the underlying data or background information.

- Avoids disconnects between the query and the contextual evidence, ensuring that the answer is not only pertinent to the question but also to the available data.

### 4.3 Faithfulness (Context-Only)

Faithfulness measures how accurately the response reflects the given context without introducing external assumptions. In this evaluation, the focus is on:

- Ensuring that all claims or conclusions in the answer are directly supported by the SQL query results.

- Avoiding any extrapolations or additions that are not justified by the context.

- Maintaining a strict adherence to the facts and figures presented in the SQL output, thereby building trust in the answer's integrity.

### 4.4 Answer Correctness (Based on Ground Truth)

This category assesses the factual accuracy of the response relative to a human curated answer—the ground truth. It examines whether the answer:

- Correctly interprets and represents the data from the SQL queries.

- Provides accurate computations, summaries, or comparisons as dictated by the ground truth.

- Is free from errors or misinterpretations, ensuring that the response can be relied upon for decision-making or further analysis.

Table 1: Judging scale

| Score | Criteria |
|---|---|
| 0 | Answer is completely wrong or does not answer the question |
| 1 | Answer contains some correct elements, but the overall response is incorrect or misleading |
| 2 | Answer contains some correct elements, but no incorrect information. The overall response may be indeterminate. |
| 3 | The answer contains all correct information, but the overall response is still indeterminate due to vagueness or lack of elaboration |
| 4 | The answer is correct but poorly formatted or expressed |
| 5 | The answer is an acceptable answer which could be obtained from another SME |

## 4.5 Evaluation results

This evaluation compares two methods—Non-Function Calling and Function Calling—using two distinct metric types: LLM-computed metrics (Answer Relevance, Relevance, Faithfulness) and human-generated correctness metrics evaluated by Subject Matter Experts (SMEs).

### 4.5.1 LLM-Computed Metrics:

Both methods performed similarly on Answer Relevance, Relevance, and Faithfulness, with the Function Calling approach slightly outperforming Non-Function Calling. Specifically, average scores were marginally higher for Function Calling across all metrics: Answer Relevance (4.55 vs. 4.50), Relevance (4.60 vs. 4.50), and Faithfulness (4.40 vs. 4.36). The distributions indicate both methods predominantly received high scores, reflecting primarily the capabilities of the underlying LLM (GPT-4o). This similarity arises because the metrics reward responses even when the agentic system fails to retrieve data if the model accurately indicates the absence of information, penalizing only explicit hallucinations. Figure 3 illustrates a comparison of average MLflow scores by approach while Figure 4 shows the distribution of MLflow scores across answer relevance, relevance, and faithfulness.

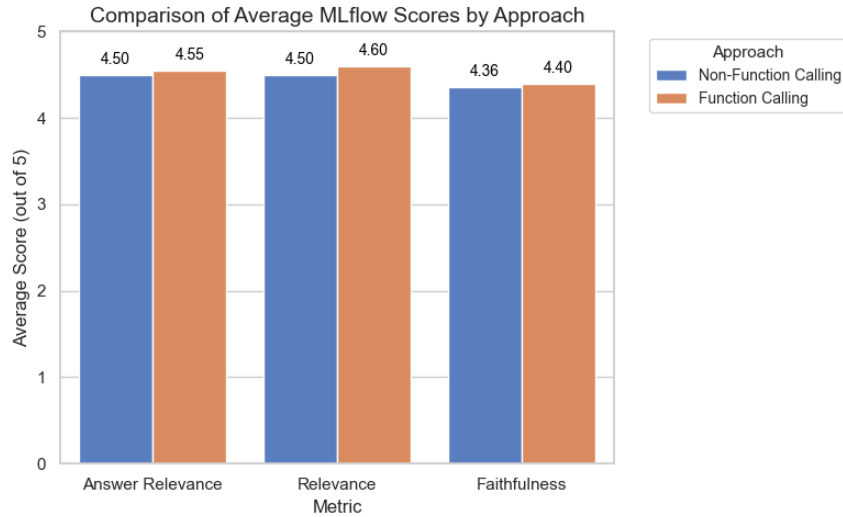

Figure 3: Comparison of average MLflow scores by approach

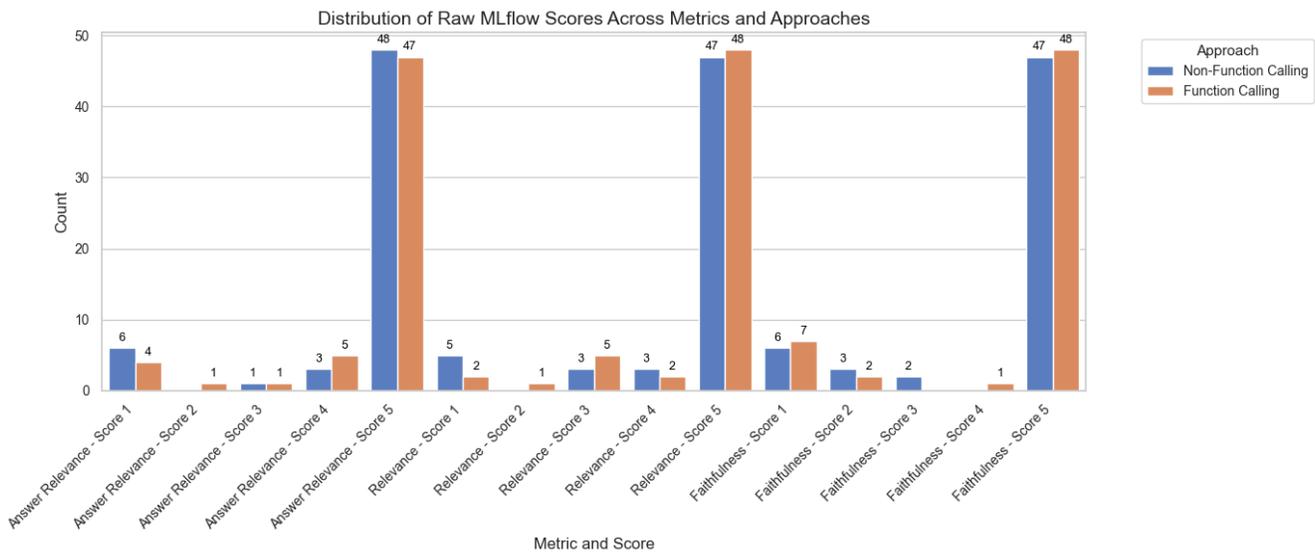

Figure 4: Distribution of MLflow scores across Answer Relevance, Relevance, and Faithfulness

### 4.5.2 Human-Evaluated Correctness Metrics:

The Function Calling method clearly outperformed the Non-Function Calling method in human-evaluated correctness scores, achieving notably higher average scores. Examining individual questions highlights that the Non-Function Calling approach frequently produced entirely incorrect answers. In contrast, the Function Calling method was consistently correct or partially correct, with fewer errors overall, due to its partially deterministic nature. Figure 5 illustrates a Comparison of average correctness scores and detailed scores per question.

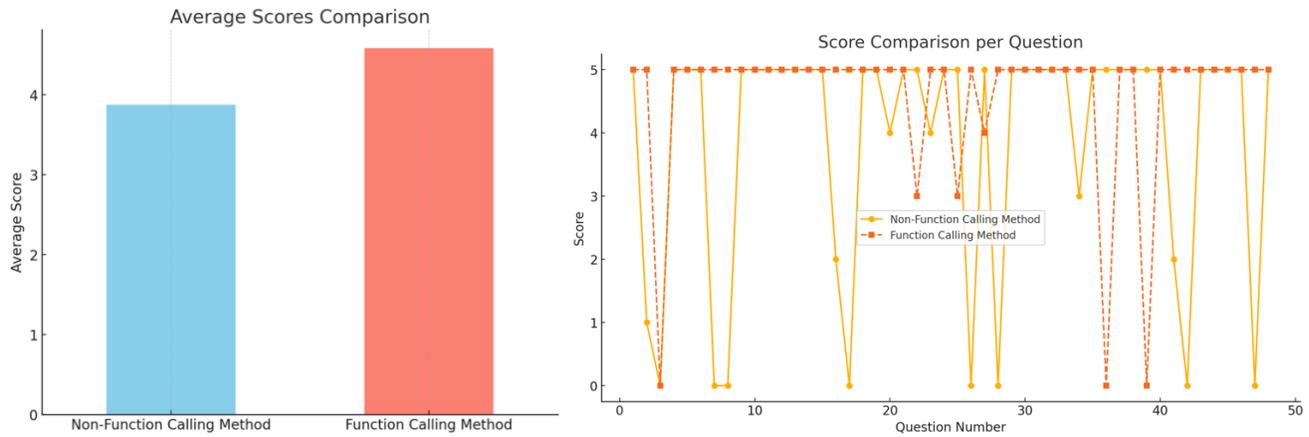

Figure 5: Comparison of average correctness scores and detailed scores per question

In summary, although LLM-computed metrics suggest only slight differences between approaches, human correctness evaluations clearly indicate a superior performance by the Function Calling approach.

## 5. Conclusion and Future Work

The proposed function-calling methodology demonstrates clear improvements in accuracy and system maintainability compared to traditional NL-to-SQL approaches. By constraining SQL generation through validated functions and structured agent workflows, the framework minimizes risks and bolsters operational safety. Future work will explore advanced reasoning models and further optimizations to ensure even more robust data retrieval in critical nuclear plant environments.

As for future work, we are considering several improvements including:

- Fine-Tuning for Domain Jargon: Develop specialized LLMs that are fine-tuned to understand and accurately interpret technical acronyms and domain-specific terminology, reducing the burden on manual prompt guidance.
- Optimized Function Selection: Fine-tune models to identify and call the correct function on the first attempt. Incorporating a custom structured output format for inter-agent communication could streamline this process and minimize unnecessary retries.
- Exploration of a Pure Function-Calling Approach: Investigate scenarios where the LLM is provided only with the database schema to answer queries. While our preliminary experiments with complex, jargon-heavy schemas were unsuccessful, this approach might be effective for databases with clearer and more standardized schemas.
- Enhanced Multi-Agent Reasoning for Chained Queries: Integrate advanced reasoning models capable of decomposing complex queries into sequential, manageable sub-tasks. This would improve the system's ability to handle chained tool calls where multiple functions must be orchestrated to retrieve the desired data.

- Dynamic Function Filtering: Implement adaptive filtering strategies—potentially leveraging retrieval-augmented generation (RAG) and reinforcement learning—to dynamically narrow down available functions based on query context, thereby reducing cognitive overload on the LLM.
- Refinement of Logging and Error Handling: Further improve error detection, retry mechanisms, and logging practices to better manage unexpected behaviours and ensure robust system performance during real-world operations.
- Investigation into Model Scalability: Evaluate the trade-offs between using large, versatile models and smaller, specialized ones that may offer cost-effective solutions while still supporting both conversational and function-calling tasks effectively.

## 6. Acknowledgments

This research paper was supported by Ontario Power Generation (OPG) and by The Natural Sciences and Engineering Research Council of Canada (NSERC) and The Canadian Nuclear Safety Commission (CNSC) grant number ALLRP 580442-2022.